\documentclass[10pt,twocolumn,letterpaper]{article}

\usepackage{cvpr}      

\definecolor{turquoise}{cmyk}{0.65,0,0.1,0.3}
\definecolor{purple}{rgb}{0.65,0,0.65}
\definecolor{dark_purple}{rgb}{0.5,0,0.5}
\definecolor{dark_green}{rgb}{0, 0.5, 0}
\definecolor{orange}{rgb}{0.8, 0.6, 0.2}
\definecolor{red}{rgb}{0.8, 0.2, 0.2}
\definecolor{darkred}{rgb}{0.6, 0.1, 0.05}
\definecolor{blueish}{rgb}{0.0, 0.3, .6}
\definecolor{light_gray}{rgb}{0.7, 0.7, .7}
\definecolor{pink}{rgb}{1, 0, 1}
\definecolor{greyblue}{rgb}{0.25, 0.25, 1}

\newcommand{\ConvexPairs}{\mathcal{C}}
\newcommand{\Similarity}{\mathrm{sim}}

\usepackage[ruled,vlined]{algorithm2e} 
\usepackage{amsfonts,amsmath,amssymb}
\usepackage{bbm}
\usepackage{amssymb}
\usepackage{float}
\usepackage{booktabs}
\usepackage{overpic}
\usepackage{color}
\usepackage{comment}
\usepackage{dsfont}
\usepackage{enumitem}
\usepackage{epsfig}
\usepackage{graphicx}
\usepackage{footnote}

\usepackage{algorithmic}
\usepackage{makecell}
\usepackage{multirow}
\usepackage[accsupp]{axessibility}
\usepackage{tabularx}
\usepackage{wrapfig}
\usepackage{xcolor}
\usepackage{xspace}

\let\originalparagraph\paragraph
\renewcommand{\paragraph}[2][.]{\vspace{-\baselineskip}\originalparagraph{#2#1}}

\newcolumntype{L}{>{\raggedright\arraybackslash}X}
\newcolumntype{C}{>{\centering\arraybackslash}X}

\makeatletter
\DeclareRobustCommand\onedot{\futurelet\@let@token\@onedot}
\def\@onedot{\ifx\@let@token.\else.\null\fi\xspace}
 
\def\ie{\emph{i.e}\onedot}

\makeatother

\definecolor{mkcolor}{RGB}{255,0,255}
\definecolor{mkbad}{RGB}{255,0,0}

\definecolor{vovacolor}{RGB}{255,0,0}

\definecolor{mhcolor}{RGB}{0,128,0}

\definecolor{nmcolor}{RGB}{0,0,255}

\definecolor{sidcolor}{RGB}{128,0,128}

\definecolor{leocolor}{RGB}{255,128,0}

\definecolor{cvprblue}{rgb}{0.21,0.49,0.74}
\usepackage[pagebackref,breaklinks,colorlinks,allcolors=cvprblue]{hyperref}

\def\paperID{-} 
\def\confName{CVPR}
\def\confYear{2026}

\title{Learning Convex Decomposition via Feature Fields}

\author{
\hspace{-2em}
Yuezhi Yang$^{1,2}$\thanks{Work done while interning at NVIDIA.} \quad
Qixing Huang$^{2}$ \quad
Mikaela Angelina Uy$^{1}$\thanks{Equal contribution.} \quad
Nicholas Sharp$^{1}$\footnotemark[2] \\
{
\textsuperscript{1}NVIDIA
\quad
\textsuperscript{2}The University of Texas at Austin
}
}

\begin{document}
\maketitle
\begin{abstract}

This work proposes  a new formulation to the long-standing problem of convex decomposition through learning feature fields, enabling the first feed-forward model for open-world convex decomposition. 
Our method produces high-quality decompositions of 3D shapes into a union of convex bodies, which are essential to accelerate collision detection in physical simulation, amongst many other applications.
The key insight is to adopt a feature learning approach and learn a continuous feature field that can later be clustered to yield a good convex decomposition via our self-supervised, purely-geometric objective derived from the classical definition of convexity.
Our formulation can be used for single shape optimization, but more importantly, feature prediction unlocks scalable, self-supervised learning on large datasets resulting in the first learned open-world model for convex decomposition.
Experiments show that our decompositions are higher-quality than alternatives and generalize across open-world objects as well as across representations to meshes, CAD models, and even Gaussian splats. 
\url{https://research.nvidia.com/labs/sil/projects/learning-convex-decomp/}

\end{abstract}
    
\vspace*{-1.5em}
\section{Introduction}
\label{sec:intro}

Convex decomposition approximates detailed, nonconvex 3D shapes with a simple set of convex bodies---these convex approximations are essential geometric accelerations for fast collision detection~\cite{Bergen1999AFA, 2083, 10.1145/285857.285860}, signed-distance computation~\cite{liu2023marchingprimitives}, motion animation ~\cite{10.1145/2461912.2461934,10.1145/1128888.1128919}, and more, which all must efficiently compute distances and spatial bounds.
Traditionally, such decompositions were manually authored by technical artists as part of the content creation process, but the prevalence of automated 3D pipelines and generative AI has created ever-increasing demand for algorithmic construction of convex decompositions.
In particular, recent progress in physical robotics training environments demands robust, high-performance simulation of generated assets and world models---convex decompositions are essential to enable these simulations~\cite{andrews2024navigation}.

\begin{figure}[t]
\vspace{-1em}
\centering
\includegraphics{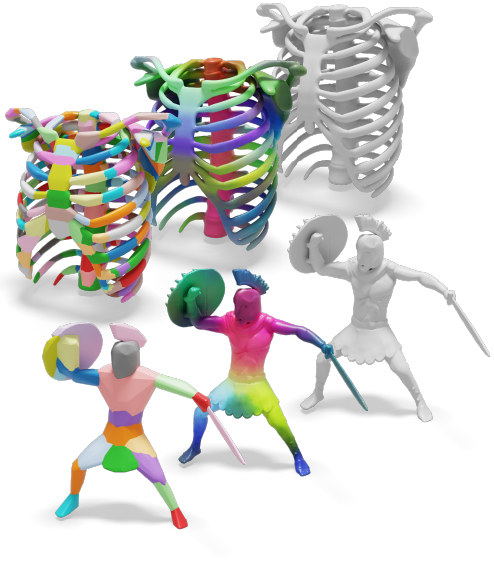}
\vspace*{-2em}
\caption{Our method takes an input shape (\emph{top}), infers features from an open-world model learned with our new self-supervised geometric loss (\emph{middle}), and clusters those features to fit the shape with a collection of tight convex bounding proxies (\emph{bottom}).
\vspace*{-1em}
}
\label{fig:teaser}
\end{figure}

However, convex decomposition has thus far remained a remarkably challenging algorithmic task; one must accurately approximate the starting shape in the geometric sense, as well as a combinatorial covering problem that is formally NP-hard in the worst case~\cite{chazelle1997strategies, orourke1983nphard}.
Traditional approaches from computational geometry use branch-and-bound techniques to explore a search space of possible decompositions, but can be prohibitively computationally expensive~\cite{doi:10.1137/0213031, 10.1145/237218.237246, lien2007acd}.
The learned architectures have also explored representing shapes with convex primitives, but thus far have been limited to narrow families of objects rather than to general open-world content~\cite{chen2020bspnet, deng2020cvxnet}.
Moreover, much past work has focused almost entirely on mesh decomposition, whereas geometry increasingly comes from imprecise representations such as Gaussian splats.

This work proposes a new formulation for convex decomposition, which allows us to train a feedforward open-world model directly producing high-quality convex decompositions of shapes.
Our key insight is to adopt a \emph{feature learning} approach.
Rather than directly optimizing or learning a discrete set of primitives, we instead operate on continuous features defined along the shape, constructing a set of features such that clustering them yields a good convex decomposition.
To make this possible, we introduce a new self-supervised contrastive loss on features, inspired by a classic geometric definition of convexity: lines connecting pairs of points should be contained within the shape.

\begin{figure}[b]
  \centering

   \includegraphics[width=1.0\linewidth]{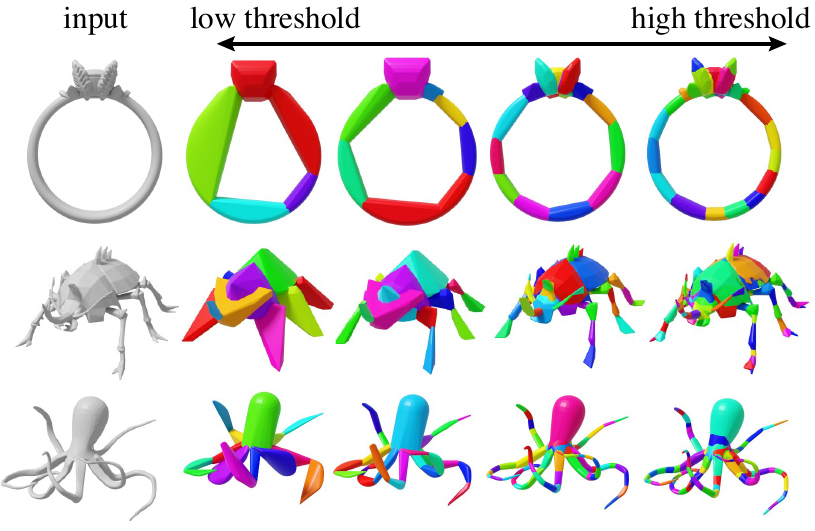}
   \caption{
   By adjusting the clustering threshold, our method can generate decompositions at varying granularity, all from the same feature field.
   }
   \label{fig:granularity}
\end{figure}

We validate our approach against both classical and deep learning baselines, demonstrating superior performance across multiple datasets. Beyond quantitative gains, we also showcase the practical utility of our method, enabling 
collision detection, control over granularity (Fig.~\ref{fig:granularity}), and generalization to various input modalities such as Gaussian splats. 
The main contributions of this work are summarized as follows:
\begin{itemize}
    \setlength\itemsep{0.4em}
    \item  We formulate convex decomposition as a contrastive learning problem and introduce our novel, self-supervised, geometric loss that enables scalable learning of convex decomposition on open-world data.
    
    \item  We train a feedforward network that enables state-of-the-art performance on convex decomposition, fast inference, and generalization to different 3D modalities.
    
    \item We demonstrate the effectiveness of our approach in different downstream applications such as multi-granularity decompositions, collision detection and generalization to different input modalities.

\end{itemize}

\pagebreak
\section{Related Work}
\label{sec:related}

\vspace{0.5em}

\paragraph{Classic Convex Decomposition}
Convex decomposition is a long-studied problem in computational geometry~\cite{doi:10.1137/0213031,orourke1983nphard,doi:10.1137/0221025,10.1145/285857.285860} enabling applications such as fast and precise collision detection~\cite{Bergen1999AFA, 2083, 10.1145/285857.285860}, shape deformation~\cite{10.1111:j.1467-8659.2007.01058.x}, distance computation~\cite{tang2014ccd,gjk1988distance}, skeleton extraction~\cite{10.1145/1128888.1128919}, animation~\cite{10.1145/2461912.2461934}, segmentation ~\cite{gadelha2020label}, simulation and gaming~\cite{andrews2024navigation}. The task of \emph{exact convex decomposition}—decomposing a shape into the minimum number of strictly convex components—is known to be NP-hard~\cite{chazelle1997strategies,orourke1983nphard}. Multiple works~\cite{doi:10.1137/0221025,10.1145/237218.237246,doi:10.1137/0213031,HERSHBERGER1998129, https://doi.org/10.1002/nme.1620370409} introduce different heuristics to tackle this problem but often produce a large number of components that limit practical use. This led to the problem of \emph{approximate convex decomposition}~\cite{lien2007acd} that instead only requires components to be \emph{nearly} convex, where they first define a concavity metric that measures the deviation of a component with its convex hull, then iteratively decomposes a shape until the concavity metric for all components is below a selected threshold. Concavity metrics have been defined based on the boundary of the component~\cite{lien2007acd, ghosh13, liu16, hacd}, volume~\cite{vhacd, attene08, 10.1145/3272127.3275029} or surface visibility~\cite{5540225, 6126256} with respect to its corresponding convex hull. Notably, V-HACD~\cite{vhacd} has long been widely-used due to easily-available implementations and robust performance, while recently CoACD~\cite{wei2022approximate} improved it by introducing a collision-aware concavity metric based on both the shape's boundary and interior volume. RL-ACD ~\cite{RL-ACD} replaces searching with reinforcement learning to improve efficiency. Although these methods perform acceptably well on general shapes, the search space over all partitions to minimize concavity is enormously large, resulting in methods that are computationally slow even with assumptions such as axis-aligned cuts~\cite{wei2022approximate} and voxelization~\cite{vhacd} to remain tractable.

\paragraph{Learning-based Methods} The prevalence of deep learning has spurred efforts to replace these classical search algorithms with data-driven optimization. Lacking ground-truth labels for optimal convex decompositions, learning-based methods such as BSP-Net~\cite{chen2020bspnet} and Cvx-Net~\cite{deng2020cvxnet}, adopt a self-supervised paradigm, where they define a fixed set of half-planes, and learn to optimize and assemble them to reconstruct the target shape. Although these approaches have shown promise in category-level datasets~\cite{chang2015shapenet}, the reconstruction objective fundamentally limits their scalability to large-scale topological and geometric variations in open-world data~\cite{objaverseXL, objaverse}. Related to this line of work is learning-based shape abstraction, where the goal is to approximate the input shape into simple primitives. These works are commonly self-supervised with reconstruction~\cite{abstractionTulsiani17, Paschalidou2019CVPR, ganeshan2025resfit, sharma2022prifit} or rendering~\cite{gao2025selfsupervisedlearninghybridpartaware, monnier2023dbw} to abstract the shape into simple primitives such as cuboids~\cite{zou20173d, abstractionTulsiani17, 10.1145/3450626.3459873, 10.1145/3355089.3356529}, superquadrics~\cite{Paschalidou2019CVPR, liu2023marchingprimitives, fedele2025superdec, gao2025selfsupervisedlearninghybridpartaware,wang2025lightsq}, superellipses~\cite{zhao2024sweepnet} or differentiable support functions~\cite{dsf}.  Part-based decomposition~\cite{liu2025partfield, Tertikas2023CVPR, Paschalidou2020CVPR,yang2025genanalysis} has also been explored and shares some similarities with our setting, however, it does not result in a good solution for convex decomposition.

\section{Method}
\begin{figure*}[t]
  \centering
   \includegraphics{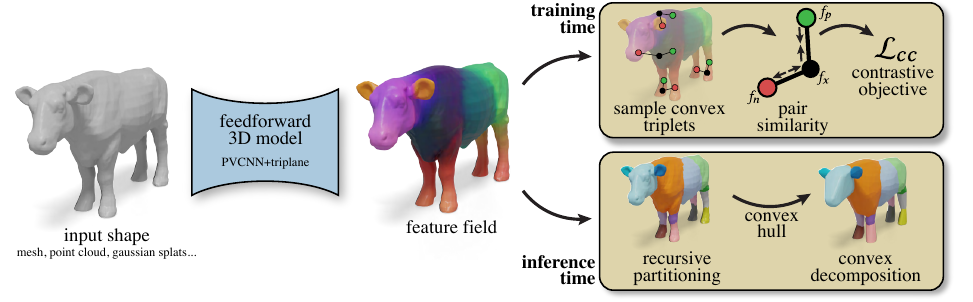}
   \caption{An overview of our convex decomposition pipeline. 
   We train a feedforward model that takes a point-sampled 3D shape as input and predicts a feature field represented defined over the object. 
   At training time, these features are fit with a self-supervised geometric objective derived from the definition of convexity.
   At inference time, the features are clustered to split the shape into components, and the convex hull of each component becomes the decomposition. Note that feature colors are visualized by running PCA on the feature field.
   \vspace*{-1em}
   } 
   \label{fig:pipeline}
\end{figure*}

Convex decomposition takes a 3D shape as input and outputs a collection of convex bodies tightly approximating the shape.
The core of our approach is to generate a set of features on the surface of the shape such that the distance in the feature space indicates points which should lie in the same convex body.
These features are fitted using a new self-supervised geometric loss (Sec~\ref{sec:contrastive_feature_learning}), and are ultimately be learned with a feedforward model (Sec~\ref{sec:method:model}).
Any application-appropriate clustering-like strategy could be applied to these features; we propose a simple recursive algorithm well-suited to this setting (Sec \ref{sec:method:outterloop}) to partition the surface, and finally evaluate the convex hull of each partition as the resulting decomposition. Figure~\ref{fig:pipeline} shows an overview of our pipeline.

\subsection{Formulation}

We introduce our approach abstractly on a general shape  $\mathcal{M}$, and will later demonstrate how to apply it to various shape representations like meshes, point clouds, and Gaussian splats.
Specifically, let $\mathcal{M} \subset \mathbb{R}^3$ denote the shape surface that we assume to be the boundary of a solid, and $\mathrm{Vol}(\mathcal{M})\subset \mathbb{R}^3$ as the solid volume. 

Convex decomposition can be viewed as finding a partition of $\mathcal{M}$ into a set of disjoint components $\{S_i\}$, that is $\bigcup_i S_i=\mathcal{M}$ and $S_i\cap S_j=\emptyset \text{ for } i\neq j$.
A good partition is one where (i) the number of components is minimized, and (ii) each component is a tight convex approximation of the shape, \ie{}
the deviation of $S_i$ from its convex hull $\mathrm{hull}(S_i)$ is small. 
We call the latter measure \emph{concavity}, and in Sec.~\ref{sec:evaluation:comparison} we will discuss exactly how it is evaluated.
We can also define an assignment function $G:\mathcal{M} \to \{S_i\}$ that takes a point on the shape and returns which segment it is a member of.

\setlength{\columnsep}{1em}
\begin{wrapfigure}{r}{100pt}
    \centering
    \vspace{-1.8\baselineskip}
    \includegraphics{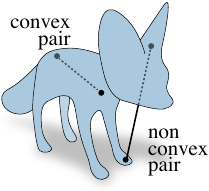}
    \label{fig:convex_pair_inset}
    \vspace{-\baselineskip}
\end{wrapfigure}
\paragraph{Convex Pairs}
Our method is inspired by a classic geometric definition of convexity.
A shape is convex if for any two points, the line segment connecting them is entirely contained inside the shape
\begin{equation}
\lambda x + (1-\lambda)y \in \mathrm{Vol}(\mathcal{M}) \quad
\forall x,y \in \mathcal{M}, \quad
\forall \lambda \in [0,1]
\end{equation}
we refer to such points as a \emph{convex pair}.
On a potentially-nonconvex shape we can define set of all convex pairs of points $\ConvexPairs(\mathcal{M}) \subseteq \mathcal{M} \times \mathcal{M}$ as
\begin{equation}
\begin{split}
\ConvexPairs(\mathcal{M}) &= \{(x \in \mathcal{M},y \in \mathcal{M} ) : \\
&\lambda x + (1-\lambda)y \in \mathrm{Vol}(\mathcal{M}) \quad \forall \lambda \in [0,1]\}
\end{split}
\end{equation}
noting that it is sufficient to test only points which lie on the surface.
Any pairs $(x,y) \notin \ConvexPairs$ are \emph{nonconvex}, meaning the line segment connecting them passes outside the shape.
In practice, a pair of points on a surface can be efficiently tested for convexity by casting a ray between its endpoints and checking for intersections with the surface.

This notion of convex pairs allows a formalization of the optimization problem for what it means to be a good, tightly-convex decomposition.
Good decompositions maximize the number of convex pairs which belong to the same segment, or equivalently minimize the number of convex pairs split into different segments
\begin{equation}
\underbrace{\vphantom{\int_{\ConvexPairs(\mathcal{M})}}\max_{\{S_i\}}}_{\substack{\textrm{optimize over}\\\textrm{partitions}}} \ 
\underbrace{\iint_{x,y \in \ConvexPairs(\mathcal{M})}}_{\substack{\textrm{for all convex pairs}\\\textrm{of points on the shape}}} \ 
\underbrace{\vphantom{\int_{\ConvexPairs(\mathcal{M})}}\mathbf{1}_{G(x) = G(y)}\, dx\, dy}_{\substack{\textrm{counting pairs which are}\\\textrm{in the same component}}} \quad
\label{eq:main_obj}
\end{equation}
\noindent where $\mathbf{1}_{G(x) = G(y)}$ is an indicator function that equal to $1$ if $x$,$y$ belongs to the same component. 
Directly optimizing this objective would amount to an intractable combinatorial partitioning problem, but we will show how to relax it to a continuous feature embedding problem.

\subsection{Convex Decomposition as Feature Learning}
We observe that the objective in Eq.~\ref{eq:main_obj} shares a similarity with unsupervised clustering, and draw inspiration from this insight to reformulate the original convex decomposition objective as a \emph{feature learning} problem.
Here, the goal is to learn continuous features that can later be clustered to yield the desired convex decomposition. 
We consider a field of $k$-dimensional features defined at each point on the shape $f : \mathcal{M} \to \mathbb{R}^k$, with some notion of feature distance $d(f_x,f_y)$, moving the argument of $f(\cdot)$ to a subscript for brevity.
Under this perspective, the objective from Eq.~\ref{eq:main_obj} can be relaxed as
\begin{equation}
\underbrace{\vphantom{\int_{\ConvexPairs(\mathcal{M})}}\min_{f}}_{\substack{\textrm{optimize over}\\\textrm{features}}} \ 
\underbrace{\iint_{x,y \in \ConvexPairs(\mathcal{M})}}_{\substack{\textrm{for all convex pairs}\\\textrm{of points on the shape}}} \ 
\underbrace{\vphantom{\int_{\ConvexPairs(\mathcal{M})}}d\big(f_x, f_y\big)\, dx\, dy}_{\substack{\textrm{distance between}\\\textrm{their features}}} \quad
\label{eq:main_obj_cont}
\end{equation}
where $f$ is the desired feature field on the shape.
As-written, Eq.~\ref{eq:main_obj_cont} tries to pull pairs of points together, and thus has a trivial degenerate solution with $f(x) = \textit{constant}$, so we balance the objective with a second term that tries to push nonconvex pairs apart

\begin{equation} \min_{f} \iint_{x,y \in \ConvexPairs(\mathcal{M})} d\big(f_x, f_y\big)\ - \iint_{x,y \notin \ConvexPairs(\mathcal{M})} d\big(f_x, f_y\big) \label{eq:contrastive_opt_plain} \end{equation}

which gives our feature embedding objective, restricting $||f|| = 1$ to prevent unbounded growth.
From a machine learning viewpoint, this objective is \emph{self-supervised}, it allows us to optimize for a good set of features, and hence a good decomposition, purely from the geometry of the input shape.
In practice, optimization amounts to first sampling many pairs $x,y$ of points on a shape, geometrically testing whether each pair $x,y \in \ConvexPairs(\mathcal{M})$ based on whether the line between the points is contained in the shape, and finally continuously optimizing for features to minimize the embedding objective.

\subsection{Contrastive Feature Learning}
\label{sec:contrastive_feature_learning}
Eq.~\ref{eq:contrastive_opt_plain} is already a well-posed embedding problem, but rather than optimizing it directly we follow previous works~\cite{chen2020simclr,liu2025partfield} to recast it as relative \emph{contrastive learning}, which is more amenable to high-dimensional stochastic optimization without imposing a metric structure.
The contrastive objective is defined by gathering triplets of points $x,p,n \in \mathcal{M}$ forming a positive pair $x,p \in \ConvexPairs(\mathcal{M})$ and negative pair  $x,n \notin \ConvexPairs(\mathcal{M})$, with the goal that the distance between the positive pair should be smaller than the distance between the negative pair.
The loss is then a log-normalized balance 
\begin{equation}
\begin{aligned}
\mathcal{L}_{\text{cc}} = -\tfrac{1}{2}\Bigg[
&\log
\frac{\Similarity(f_x, f_p)}
{\Similarity(f_x, f_p) + \Similarity(f_x, f_n)} \\
&+
\log
\frac{ \Similarity(f_p, f_x)}
{\Similarity(f_p, f_x) + \Similarity(f_p, f_n)}
\Bigg],
\end{aligned}
\label{eq:triplet_ctr}
\end{equation}
where we also exchange smaller-is-closer distance for larger-is-closer similarity, typically the exponential of cosine distance on sphere-normalized features $\Similarity(x, y) = \exp\!\big(x\cdot y / \tau\big)$, with $\tau$ as a temperature hyperparameter akin to a softmax.

\begin{figure}[b]
\centering
\includegraphics{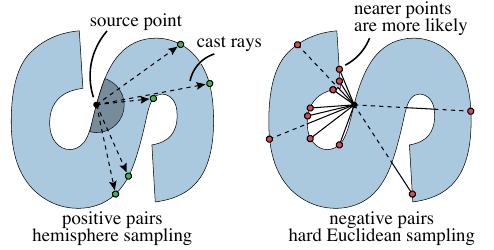}
\caption{
    Contrastive training triplets are formed by a source point, a positive pair generated by casting a ray in a random inward direction (\emph{left}), and a negative pair rejection-sampled from all points on the surface weighted to prefer nearby points (\emph{right}).
}
\label{fig:ray_sampling}
\end{figure}

\paragraph{Triplet Sampling and Hard Negatives}
The triplets $(x,p,n)$ could be generated by uniform rejection sampling, but this is computationally expensive to begin with, and furthermore there is opportunity to seek out \emph{hard negative} samples which are particularly informative to the optimization process.
We first choose an anchoring sample $x \in \mathcal{M}$ uniformly from the object's surface.
To get a sample $p \in \mathcal{M}$ forming a positive pair with $x$, we cast a random ray into the hemisphere opposite the surface-normal direction of $x$, into the interior of the shape, and takes the point where it exits the surface as $p$.
To get a sample $n \in \mathcal{M}$ forming a negative pair with $x$, we rejection sample by generating candidates on the surface of $\mathcal{M}$ and testing whether the line segment connecting $x$ and $n$ exits the shape.
Rather than uniformly gathering points on the surface, we prefer negatives spatially close to $x$, sampling inversely proportional to the Euclidean distance $\mathrm{P}(n) = \frac{1}{\|n - x\|^2}$.
These negatives are likely to be challenging pairs, enabling more efficient optimization of features.

The geometric queries in this sampling procedure can be implemented efficiently and robustly, as it requires only sampling points and casting rays, the latter of which
can be hardware-accelerated with libraries like Intel Embree and NVIDIA OptiX for fast on-the-fly triplet generation.

\begin{figure}
\centering
\begin{overpic}[width=\linewidth]{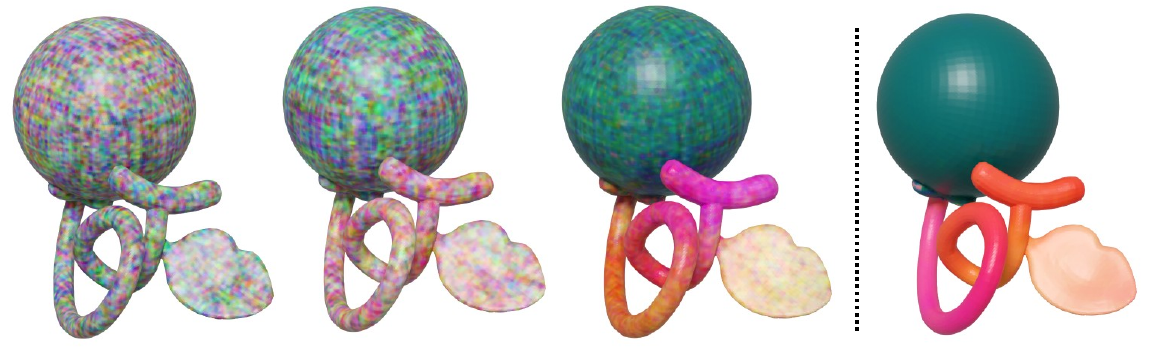}
    \put(15,-2){\small Single shape optimization process}
    \put(76,-2){\small Feedforward}
\end{overpic}

\caption{Left: Visualization of the learned feature field during our single-shape optimization, shown at different optimization stages. Right: Feature fields produced directly by our feed-forward model.}
\label{fig:single_optimization}
\end{figure}

\paragraph{Single shape optimization}
Given the sampling and optimization procedures described above, we can optimize our feature fields directly on a single shape to obtain its convex decomposition. 
Figure~\ref{fig:single_optimization} shows an example of such optimization. 
Although already effective, we take advantage of feature-based setting to train a feedforward model to generate smoother features, rather than fitting them per-shape.

\subsection{Feedforward Model}
\label{sec:method:model}
Feature-based formulations have been used throughout visual computing as a framework for large-scale learning of segmentation, embeddings, and more~\cite{caron2021dino,oquab2023dinov2,radford2021clip}.
Reformulating convex decomposition as a feature learning problem makes it amenable to this approach, allowing us to train a feedforward model predicting the field $f$ conditioned on an input shape $\mathcal{M}$, self-supervised across a large dataset of open-world 3D shapes~\cite{objaverse, objaverseXL}. 
Our self-supervised geometric loss (Eq.~\ref{eq:triplet_ctr}) is essential, sidestepping the lack of high-quality ground truth supervision for this task to learn solely from the shapes' geometry.
Compared with per-shape optimization, the feedforward model offers three main advantages: (a) fast inference, (b) smooth feature fields that are robust to input noise or incomplete geometry (c) generalization across 3D input modalities, i.e. our model can be applied to different 3D representations at inference time, without requiring a watertight mesh.

We adopt an architecture similar to prior work on open world-shape learning~\cite{liu2025partfield}. 
The model takes as input a point cloud sampled from the shape's surface $\mathcal{M}$, and outputs a triplane-encoded feature field that can be evaluated at any spatial location. 
This network consists of two main stages. 
First, a PVCNN encoder~\cite{liu2019pvcnn} encodes an input shape $\mathcal{M}$ represented as a point cloud and extracts per-point features from the input. 
These features are orthogonally projected onto three axis-aligned 2D feature planes via mean reduction 
to form an initial triplane representation. 
These initial triplanes are processed by a 2D CNN for downsampling, then reshaped and passed through a transformer module, and finally upsampled via a transposed 2D CNN to reconstruct final feature triplanes. Hence  for any 3D query point we have its corresponding feature is retrieved by aggregating corresponding features from the final feature triplane.

\vspace{-2mm}

\subsection{Recursive Decomposition on Features}
\label{sec:method:outterloop}
\vspace{-2mm}

At inference time, our trained model generates feature fields $f: \mathcal{M} \to \mathbb{R}^k$, such that similar features indicate surface regions that should appear together in the decomposition.
Concretely, we densely sample features on the surface of the input shape $S$ using the predicted field and apply off-the-shelf clustering algorithms to partition $S$ into approximate convex components $\{S_i\}$.
In principle any clustering algorithm could be used for this purpose, common choices include $k$-means for fast performance, or agglomerative clustering to take connectivity into account.
For mesh-based inputs, features are sampled per face and clustered using agglomerative clustering with mesh connectivity. 
For other modalities such as point clouds, we blend feature-space distance and Euclidean distance and apply $k$-means clustering.
Finally, we compute convex hulls $\{\mathrm{hull}(S_i)\}$ for each cluster, whose union approximates the input geometry. 

\begin{figure}
\centering
\setlength{\abovecaptionskip}{2pt}
\begin{overpic}[width=0.9\linewidth]{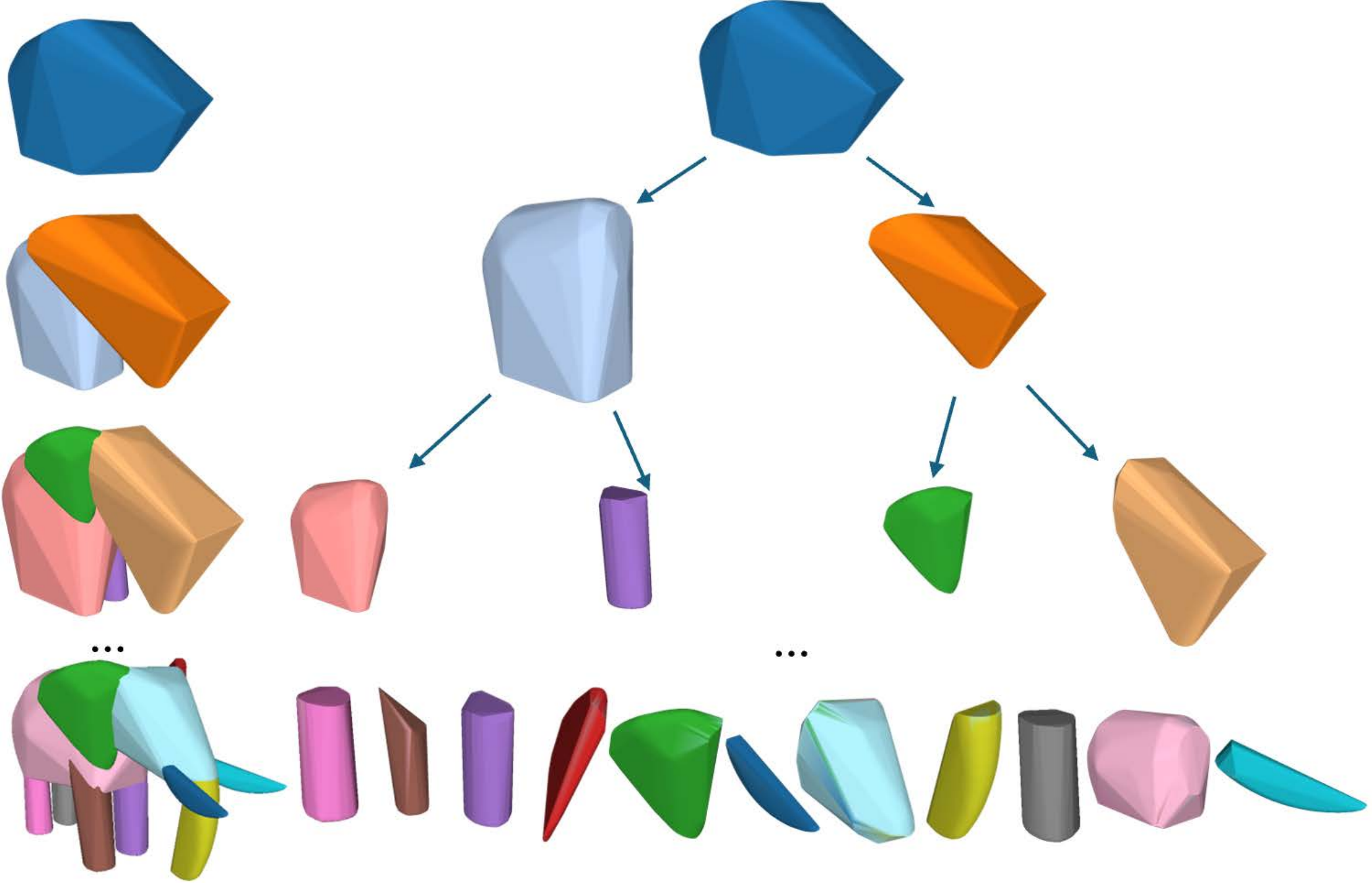}
    \put(35,65){\small{Decomposition Hierarchy}}
    \put(-1,65){\small{Convex Union}}
\end{overpic}
\caption{A hierarchical decomposition tree from our algorithm.}
\label{fig:binary_tree}
\vspace*{-1em}
\end{figure}

There is no obvious strategy to choose the number of clusters in the decomposition of a shape, so we adopt a recursive strategy of repeatedly binary clustering until a desired user-specified concavity threshold is reached.
This threshold allows control over the granularity of the output; importantly this granularity need only be set during the clustering post-processing, and our learned features can be used for decomposition at any granularity.
Algorithm~\ref{tab:recur_algo} describes the divide-and-conquer strategy. For each cluster which is not yet a sufficiently tight convex proxy for its covered region, we apply binary clustering to split it, then recurse on both subcomponents. 
Components are processed in order of concavity until all reach the target threshold, or a user-specified maximum component count is reached. Figure~\ref{fig:binary_tree} illustrates an example of this process.

\begin{algorithm}[t]
\caption{Our Recursive Decomposition}
\begin{algorithmic}[1]
\REQUIRE Mesh $\mathcal{S}$, threshold $\varepsilon$, max hulls $K$
\ENSURE Parts $\mathcal{R}$
\STATE $P\leftarrow\textsc{Sample}(\mathcal{S});\ F\leftarrow\textsc{Model}(P)$
\STATE $C_{\mathcal{S}}\leftarrow\textsc{Concavity}(\mathcal{S})$
\STATE $\mathcal{Q}\leftarrow\{(\mathcal{S},C_{\mathcal{S}})\}$ \COMMENT{max-heap by concavity}
\STATE $\mathcal{R}\leftarrow\emptyset$; $N\leftarrow0$
\WHILE{$\mathcal{Q}\neq\emptyset$ \textbf{and} $N<K$}
  \STATE $(P,C_P)\leftarrow\textsc{PopMax}(\mathcal{Q})$
  \IF{$C_P<\varepsilon$}
    \STATE $\mathcal{R}.\textsc{Add}(P)$; $N\leftarrow N+1$
  \ELSE
    \STATE $(P_1,P_2)\leftarrow\textsc{Clustering}(P,2)$
    \STATE $C_1\leftarrow\textsc{Concavity}(P_1)$; $C_2\leftarrow\textsc{Concavity}(P_2)$
    \STATE \textsc{Push}$(\mathcal{Q},(P_1,C_1))$; \textsc{Push}$(\mathcal{Q},(P_2,C_2))$
  \ENDIF
\ENDWHILE
\STATE \RETURN $\mathcal{R}$
\end{algorithmic}
\label{tab:recur_algo}
\end{algorithm}

\vspace{-2mm}

\section{Experiments}
\subsection{Implementation details}
\label{sec:evaluation:detail}
We train our model on the Objaverse dataset~\cite{objaverse}.
Low-quality data, such as scans with broken or incomplete geometry, are filtered out, leaving approximately 340K shapes for training. 
All shapes are normalized to the range $[-1, 1]^3$, and 100,000 points are uniformly sampled from each shape as network input. 
For every shape, we generate 10,000 triplets; for each triplet, 64 positive and 512 negative samples are drawn to construct candidate pairs.

The feature field has a dimensionality of 448. 
The triplane representation has a spatial resolution of $512 \times 512$ with 128 channels, and the transformer backbone consists of six layers. 
We train the model for 600,000 steps on eight NVIDIA A100 GPUs over one week, using a batch size of two per GPU. It takes 5s to generate features and 13s for the recursive algorithm to perform decomposition.

\subsection{Baseline Comparisons}
\label{sec:evaluation:comparison}
\vspace{0.8em}

\paragraph{Evaluation Datasets} 
We evaluate our method on three datasets: V-HACD~\cite{vhacd}, PartObjaverse-Tiny~\cite{yang2024sampart3d} and ShapeNet~\cite{chang2015shapenet} datasets. 
The V-HACD dataset contains 61 3D models including mechanical objects, animals, and human body parts, and is commonly used for evaluating approximate convex decomposition algorithms. 
The PartObjaverse-Tiny dataset is a curated subset of the large-scale Objaverse collection, containing 200 3D objects spanning diverse semantic categories. 
ShapeNet~\cite{chang2015shapenet} is a large dataset for man-made shapes commonly used for evaluating learning based 3D shape reconstruction. We use 13 most common categories in ShapeNet and randomly sample 10$\%$ of the shapes from the test set resulting in 900 models. 

\paragraph{Evaluation Metric} Following previous work, we evaluate all methods using the concavity metric defined in~\cite{wei2022approximate}.
We compute $\max_i (\mathrm{Concavity}(\mathcal{M} \cap \mathrm{hull}(S_i))$, where $\mathrm{Concavity}(P)$ is the maximum of the surface and volumetric Chamfer distances between $P$ and its convex hull. We also further report the overall reconstruction accuracy defined as the Chamfer distance between the input shape and the union of all convex hulls, i.e. $\mathrm{Chamfer}(\mathcal{M}, \cup_i \mathrm{hull}(S_i))$.

\paragraph{Baselines} 
We compare our method against both (i) classical convex decomposition approaches: CoACD~\cite{wei2022approximate} and V-HACD~\cite{vhacd}, as well as (ii) learning-based methods: Cvx-Net~\cite{deng2020cvxnet} and BSP-Net~\cite{chen2020bspnet}.
For fair comparison, we report both Cvx-Net and BSP-Net using their original setup trained on ShapeNet (Cvx-S, BSP-S) as well as both models trained on Objaverse, a larger dataset corpus (Cvx-O, BSP-O). We use their recommended parameter settings in all experiments.
Quantitative evaluation for CoACD and V-HACD are computed by setting hyper-parameters that maintain the same level of granularity with our method and visualize results at different granularity level qualitatively.

\begin{figure}
\centering
\setlength{\unitlength}{1pt} 
\begin{overpic}[width=\linewidth]{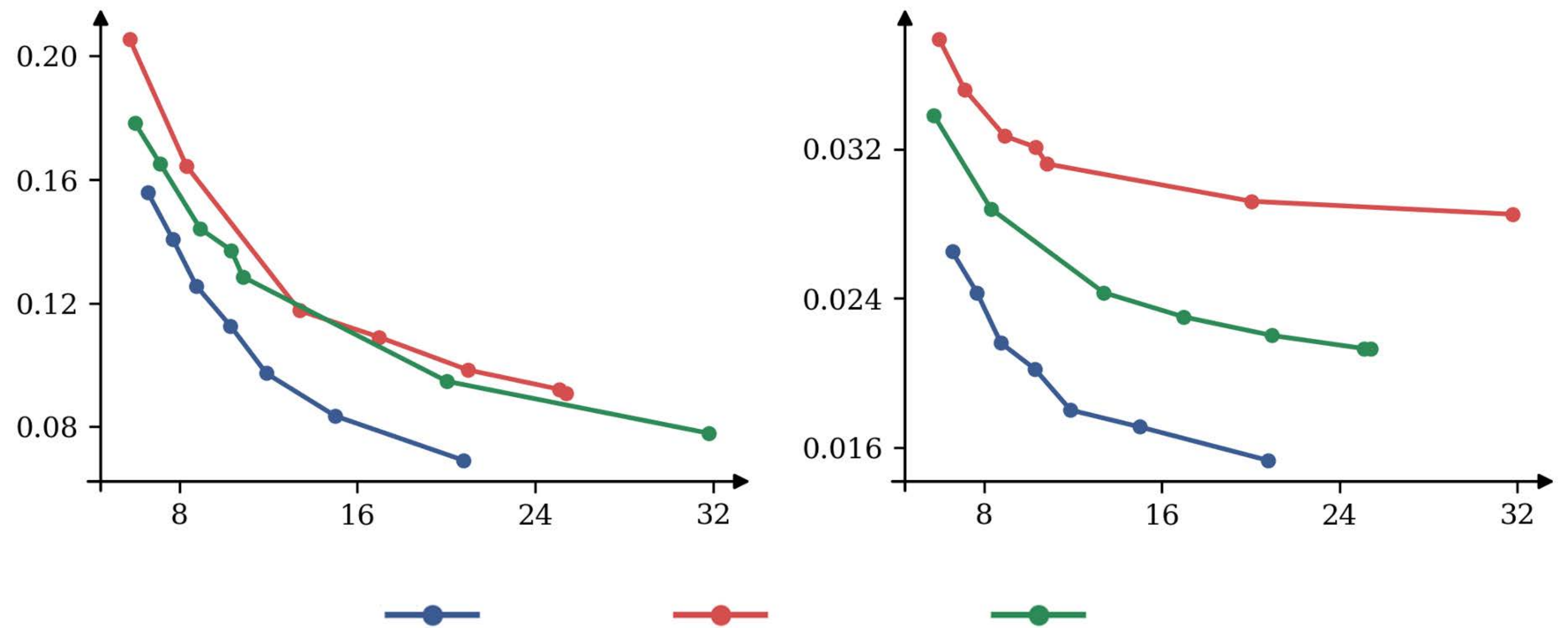}

  \put(0,40){\footnotesize Concavity $\downarrow$}
  \put(48,40){\footnotesize Recon. Error $\downarrow$}

  \put(22,4){\footnotesize \# comp.}
  \put(73,4){\footnotesize \# comp.}

  \put(31,0){\footnotesize \textbf{OURS}}
  \put(50,0){\footnotesize \textbf{VHACD}}
  \put(70,0){\footnotesize \textbf{CoACD}}
\end{overpic}
\caption{Quantitative comparison with CoACD and VHACD on the VHACD dataset under different granularity.}
\label{fig:granularity_comparison}
\end{figure}

\begin{figure*}[h]
\centering
\begin{overpic}[width=1.0\linewidth]{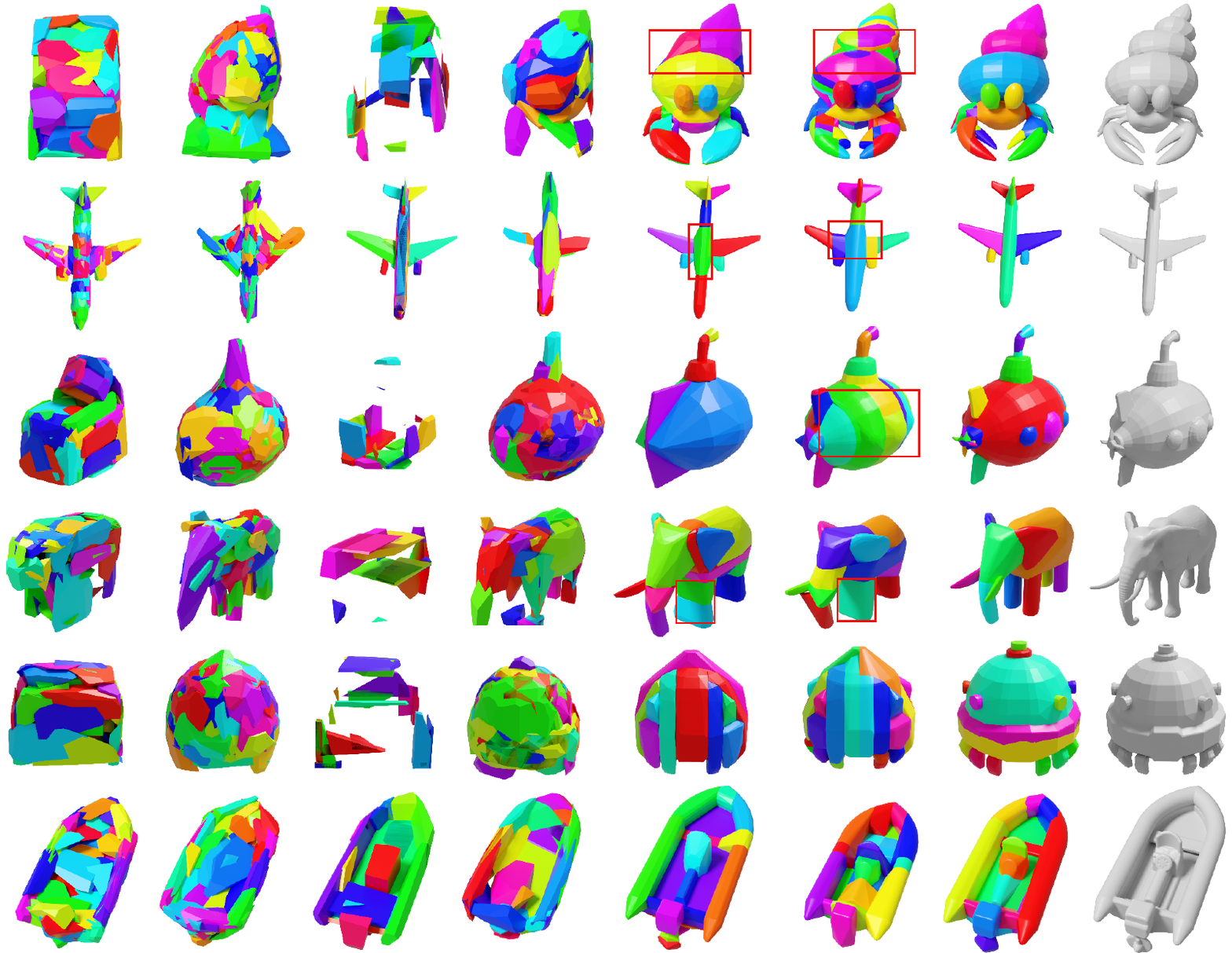}
    \put(7,71){Cvx-S}
    \put(19,71){Cvx-O}
    \put(31,71){BSP-S}
    \put(42,71){BSP-O}
    \put(53,71){VHACD}
    \put(65,71){CoACD}
    \put(77.5,71){Ours}
    \put(86,71){input shape}
\end{overpic}
\caption{Qualitative results against the baselines. 
}
\label{fig:qualitative_comparison}
\end{figure*}

\begin{table*}[t]
\centering
\begin{tabularx}{\linewidth}{c|CCC|CCC|CCC}
    \toprule
     & \multicolumn{3}{c|}{VHCD Dataset} & \multicolumn{3}{c|}{PartObjaverse-Tiny} & \multicolumn{3}{c}{ShapeNet}\\
     & \# comp.  \ & concavity & recon. & \# comp. & concavity & recon. & \# comp. & concavity & recon. \\
    \midrule
    VHCD & 13.39 & 0.1176 & 0.0213 & 21.94 & 0.1800 & 0.0318  &  14.73 & 0.1155 & 0.0201 \\
    COACD & 14.47 & 0.1095 & 0.0321 & 23.63 & 0.1388 & 0.0321 & 13.02 & 0.0746 & 0.0172 \\

    \midrule
    BSP-Net (ShapeNet) & 12.90 & 0.2249 & 0.1227 & 12.33 & 0.2593 &  0.1258& 21.58 & 0.2107 & 0.0273  \\
    BSP-Net (Objaverse) & 22.26 & 0.1857 &  0.0297&  21.44& 0.1964 & 0.0315 & 16.15 & 0.1592 & 0.0273 \\
    \midrule
    Cvx-Net (ShapeNet) & 50.00 & 0.4673 & 0.0814 & 50.00 & 0.5079 & 0.0808 & 50.00 & 0.1795 & 0.0234 \\
    Cvx-Net (Objaverse) & 50.00 & 0.2880 & 0.0373 & 50.00 & 0.2970 & 0.0359 & 50.00 & 0.2191 & 0.0258 \\
    \midrule
    \textbf{Ours} & \textbf{11.90} & \textbf{0.0973} & \textbf{0.0180} & \textbf{21.18} & \textbf{0.1257} & \textbf{0.0254} & \textbf{12.63} & \textbf{0.0656} & \textbf{0.0142} \\
    \bottomrule
\end{tabularx}
\caption{
Quantitative comparison across three datasets.
\#comp.\ denotes the number of convex components, and \emph{recon.} denotes reconstruction accuracy measured using Chamfer distance. For all metrics, lower $\downarrow$ means better.
}
\label{tab:comparison}
\end{table*}

\paragraph{Results} Quantitative results are reported in Table~\ref{tab:comparison}, where we see that across all datasets, our method consistently outperforms both classical and learning-based baselines in terms of concavity and reconstruction error at any given component count. 
Figure~\ref{fig:granularity_comparison} further comprehensively compares our method against V-HACD and CoACD showing superior performance at multiple granularity levels.
Figure~\ref{fig:qualitative_comparison} shows our qualitative results.
While these classical approaches remain strong baselines, they often fail to capture inclined convex regions, leading to unnecessary splits (top row: hermit crab shell) due to their axis-aligned cut assumption. 
In contrast, our approach better preserves large convex structures (third row: submarine), separates nearby convex parts (fourth row: elephant), and performs well even when restricted to a small number of components (second row: airplane).
On learning-based baselines, both Cvx-S and BSP-S struggle to generalize on shapes beyond those in ShapeNet as shown by their poor reconstruction results (see airplane in the second row vs other examples), even when using a substantially larger number of convexes.
While both Cvx-O and BSP-O improve in reconstruction quality, their resulting convex decomposition are still far from ideal, suggesting that both do not scale well to open-world shapes.

\subsection{Analysis and Ablation}
\vspace{0.8em}

\paragraph{Concavity thresholds $\epsilon$} Figure~\ref{fig:granularity} shows the convex decomposition produced under different user-specified concavity thresholds $\epsilon$, which directly controls the decomposition granularity. Lower thresholds produce more components and preserve finer geometric details, while higher thresholds yield fewer components and result in coarser approximations of the input shape.

\paragraph{Ablation Study} We evaluate variants of our framework to understand the impact of each component. First, we remove hard-negative sampling when selecting negative pairs and hemisphere-based sampling when selecting positive pairs in the contrastive triplet construction (Section~\ref{sec:contrastive_feature_learning}). We also compare against a version that does not use our recursive decomposition strategy (Section~\ref{sec:method:outterloop}), instead increasing the number of clusters in a flat clustering procedure until the stopping criterion is met. Finally, we include an ablation where optimization is performed directly on a single shape rather than through the feedforward model. Quantitative results for all settings are shown in Table~\ref{tab:ablation}.
\begin{table}[t]
\centering
\small
\setlength{\tabcolsep}{3.5pt}
\renewcommand{\arraystretch}{1.1}
\begin{tabularx}{\linewidth}{c|CCC|CCC}
    \toprule
     & \multicolumn{3}{c|}{VHCD Dataset} & \multicolumn{3}{c}{Objaverse-Tiny} \\
     & \#comp & conc. & recon. & \#comp & conc. & recon. \\
    \midrule

    - Hard-neg & 12.88 & 0.0993 & 0.0186 & 21.60 & 0.1293 & 0.0268 \\
    - Hemis-Pos & 13.08 & 0.0999 & 0.0219 & 23.27 & 0.1286 & 0.0263 \\
    - Recur. Clus. & 12.25 & 0.1292 & 0.0191 & 25.44 & 0.1375 & 0.0288 \\
    Single opt. & 12.40& 0.1134 & 0.0198 & 22.35 & 0.1300 & 0.0266 \\
    \textbf{Ours} & \textbf{11.90} & \textbf{0.0973} & \textbf{0.0180} & \textbf{21.18} & \textbf{0.1257} & \textbf{0.0254} \\
    \bottomrule
\end{tabularx}
\caption{Quantitative ablation on components of our approach.}
\label{tab:ablation}
\end{table}

\vspace{-5mm}
\noindent\paragraph{3D segmentation features}
Although 3D segmentation and convex decomposition share conceptual similarities, segmentation features are not suitable for convex partitioning. To highlight this difference, we apply our recursive decomposition algorithm to feature fields produced by a recent 3D segmentation model, PartField~\cite{liu2025partfield}.
PartField's features reflect semantic structure, which leads to poor convex partitions when clustering is applied.
In contrast, our method learns features that are convex-aware, enabling accurate and geometrically meaningful decompositions.

\begin{figure}
\centering
\includegraphics{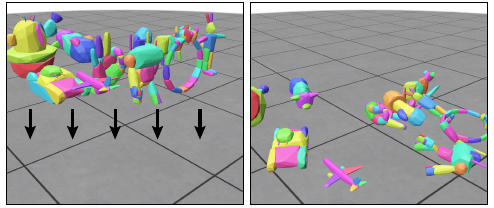}
\caption{
    Convex decompositions are used to accelerate collision handling in physical simulations, shown here on rigid bodies, decomposed with our method and simulated under gravity in the Newton engine~\cite{newton_physics_newton} .
}
\label{fig:collision}
\end{figure}

\subsection{Applications}
\label{sec:evaluation:application}
\vspace{0.8em}

\paragraph{Collision Detection}
We showcase our convex decomposition results by applying the default collision detection implementation in Newton~\cite{newton_physics_newton} as shown in Figure~\ref{fig:collision}. Our convex approximation yields a 5x faster simulation step v.s. using the original meshes (8ms vs 40ms).

\begin{figure}
\centering
\begin{overpic}[width=\linewidth]{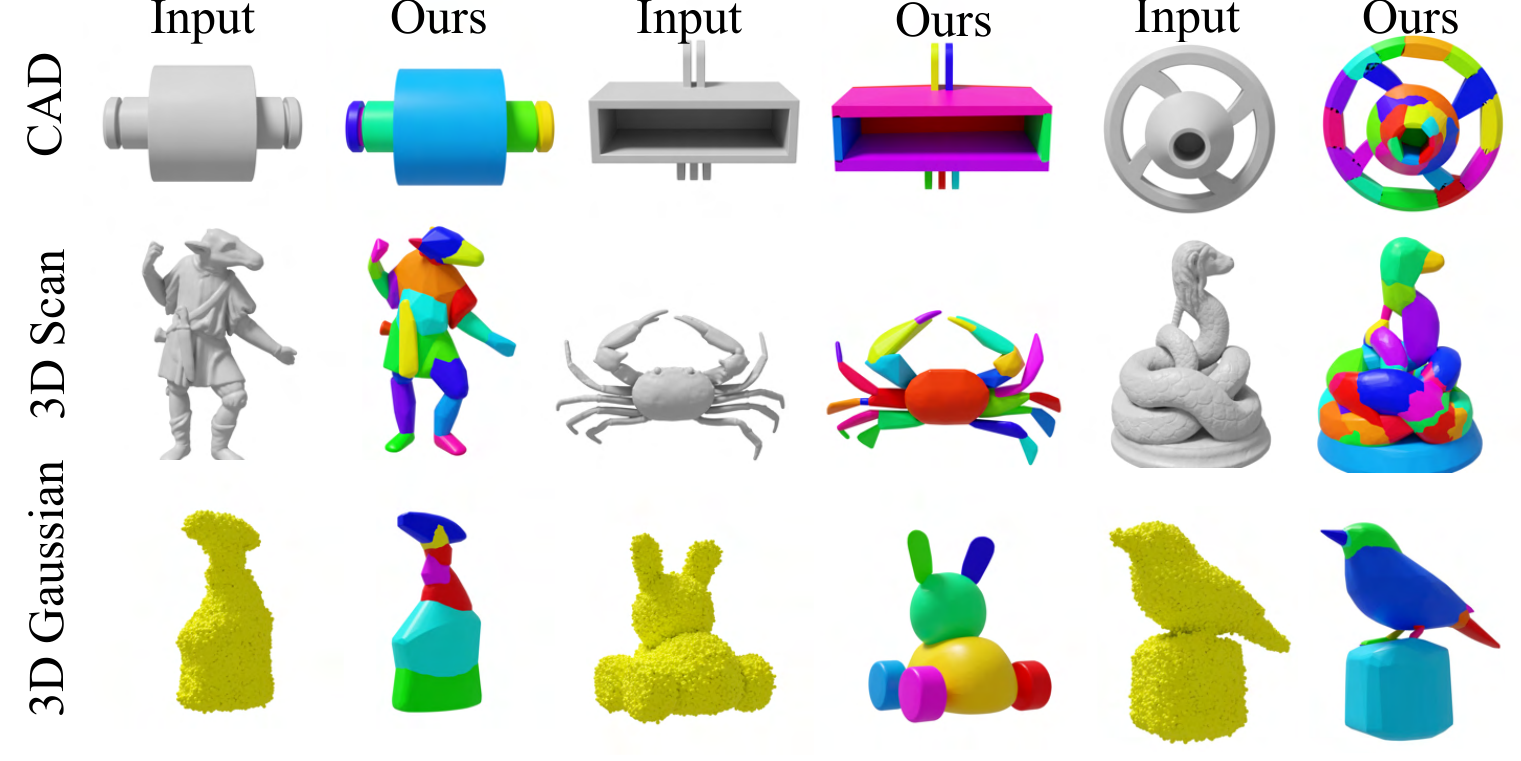}
\end{overpic}
\caption{
Our model can take different modalities as input, including CAD models, 3D scans, and 3D Gaussian splats, and perform convex decomposition.
}
\label{fig:input_modularity}
\end{figure}

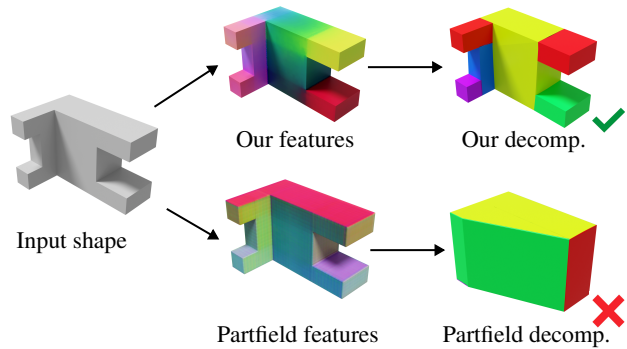
\begin{figure}
\centering
\begin{overpic}[width=\linewidth]{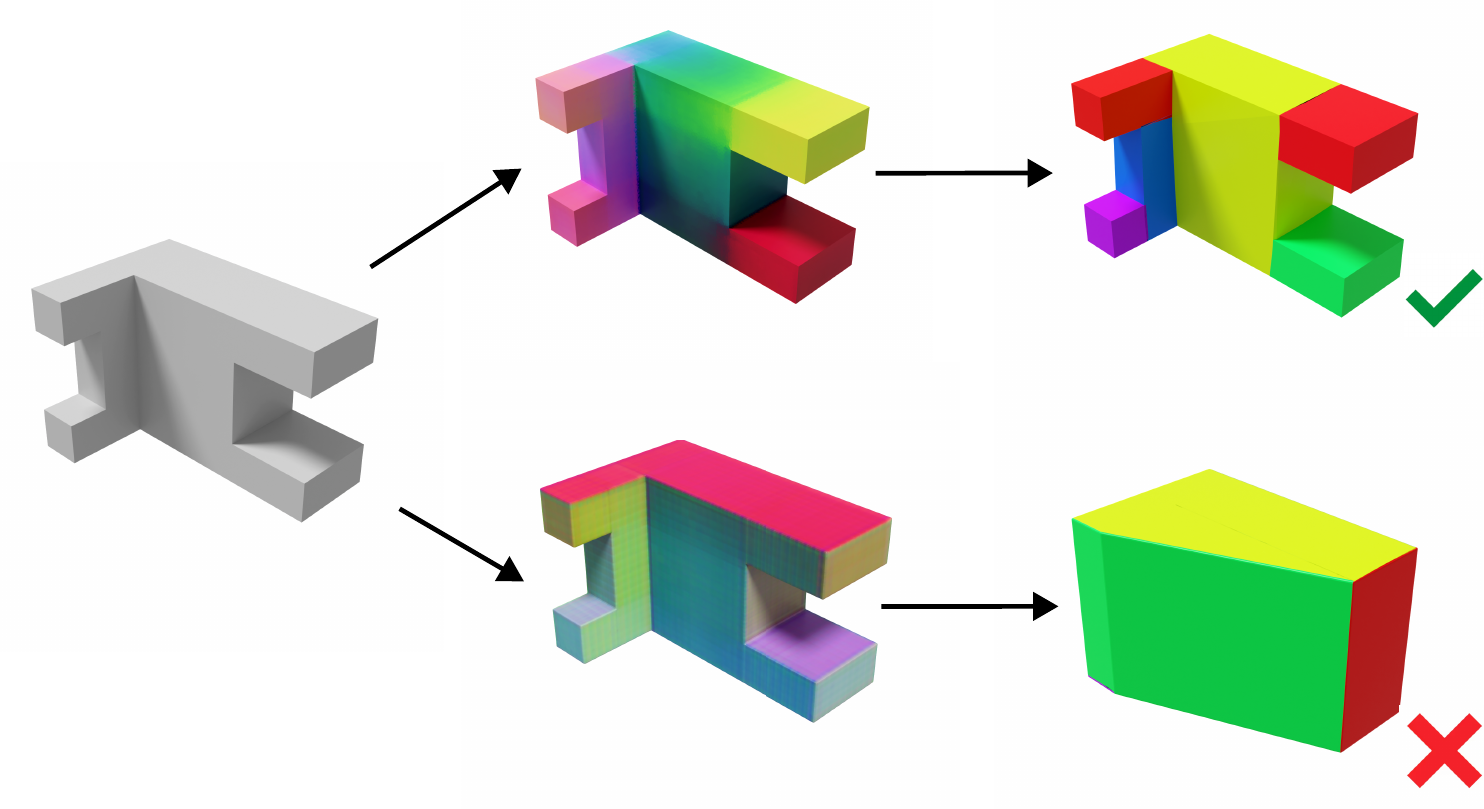}
    \put(3,14){\small Input shape}
    \put(38,30){\small Our features}
    \put(35,-1){\small Partfield features}
    \put(74,30){\small Our decomp.}
    \put(71,-1){\small Partfield decomp.}
\end{overpic}
\caption{Our geometric objective produces convex-aware features for accurate decomposition (top). PartField instead yields nearly uniform features on flat regions. For example, the L-shaped surface appears almost entirely red—resulting in inferior decomposition (bottom). Feature colors are visualized by principle component analysis on the feature field. }
\vspace{-5mm}
\label{fig:partfield_compare}
\end{figure}

\paragraph{Input Modalities}
Our feed-forward model operates directly on point clouds, allowing it to be applied to 3D shapes  in a wide range of modalities. Figure~\ref{fig:input_modularity} shows convex decompositions generated from CAD models of mechanical parts from the ABC dataset~\cite{koch2019abc}, real-world 3D scans~\cite{3dscans2025}, and AI-generated 3D Gaussian splats. Although trained primarily on point clouds sampled from human-authored meshes, the model generalizes well across these diverse input types, demonstrating broad applicability.

\section{Limitations and Future Work}
\setlength{\columnsep}{1em}
\begin{wrapfigure}[6]{r}{120pt}
    \centering
    \vspace{-1.2\baselineskip}
    \includegraphics[width=120pt]{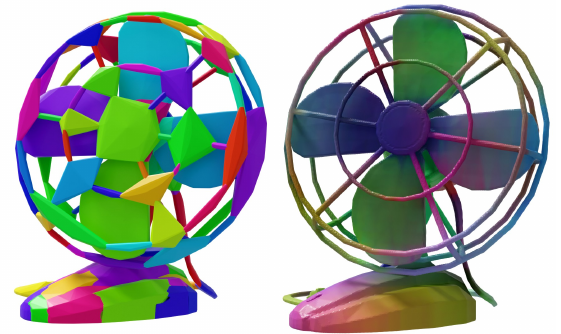}
    \vspace{-\baselineskip}
\end{wrapfigure}
Our model was trained on clean object-level data, and does not necessarily generalize to scene-scale or highly-incomplete geometry; one route to enhance this capability is training on scenes and noise-injected data.
Our model also struggles with complex thin structures such as the frame of the fan (inset).
More broadly, we are optimistic about further extending the generation of convex decompositions with feature learning, such as learning semantically-aware proxies adapted to likely motions of an object.

\newpage
\paragraph{Acknowledgments} We would like to thank Guying Lin, Anka Chen and Jun Gao for helpful discussions. Yuezhi Yang and Qixing Huang were supported by NSF-2047677, 2413161, 2504906, and 2515626; GIFTs from Adobe and Google; and computing support on the Vista GPU Cluster through the Center for Generative AI (CGAI) and the Texas Advanced Computing Center (TACC) at UT Austin.

{
    \small
    \bibliographystyle{ieeenat_fullname}
    \bibliography{main}
}

\clearpage
\setcounter{page}{1}
\maketitlesupplementary

\appendix
This document supplements the main paper \emph{Learning Convex Decomposition via Feature Fields}. In particular, we provide additional implementation details (Sec.~\ref{sec:supp:implementation}), baseline comparison details (Sec.~\ref{sec:supp:baseline_comparison}), runtime comparison with baselines (Sec.~\ref{sec:supp:run_time_comparison}), an alternative training objective ablation (Sec.~\ref{sec:supp:additional_ablation}) and more qualitative results on our proposed method (Sec.~\ref{sec:supp:more_result}).

\section{Implementation Details}
\label{sec:supp:implementation}

\subsection{Triplet Sampling}
\label{sec:supp:hardneg}
Positives and negatives for our contrastive loss are sampled on the fly during training. For each shape, we first run marching cubes on a precomputed $256^3$ signed distance function grid to obtain a watertight mesh. We then uniformly sample 1024 surface points as anchor points. For each anchor point, we cast 64 random rays into the hemisphere opposite the surface normal to obtain 64 positive candidate pairs. Negative candidates are generated via rejection sampling over the surface points of the entire shape until we obtain 1024 valid negative samples for each anchor.
Triplets are then constructed by pairing positives and negatives that share the same anchor point. Specifically, for each anchor point we sample one positive pair (x,p) uniformly from the 64 positive candidates. Each positive pair is then matched with 512 negatives $(x,\{n\})$: 256 sampled uniformly and 256 selected via hard-negative mining from the pool of 1024 negative candidates. This produces triplets 
$(x,p,\{n\})$, and the contrastive loss is defined as:
\begin{equation}
\begin{aligned}
\mathcal{L}
= -\frac{1}{2}\Bigg[
&\log\!\left(
\frac{\operatorname{sim}\!\big(f(x), f(p)\big)}
     {\operatorname{sim}\!\big(f(x), f(p)\big)
      + \sum_{n} \operatorname{sim}\!\big(f(x), f(n)\big)}
\right)
\\[6pt]
+\, &
\log\!\left(
\frac{\operatorname{sim}\!\big(f(p), f(x)\big)}
     {\operatorname{sim}\!\big(f(p), f(x)\big)
      + \sum_{n} \operatorname{sim}\!\big(f(p), f(n)\big)}
\right)
\Bigg]
\end{aligned}
\end{equation}
We utilize Embree~\cite{wald2014embree} via pyembree to accelerate ray–mesh intersection.

\subsection{Input Preprocessing and Clustering}
\label{sec:supp:inference}

For all experiments except those using 3D Gaussian splats as input, we preprocess each shape into a manifold, watertight mesh following the protocol of \cite{wang2022mesh2sdf}. Clustering is then performed using agglomerative clustering from SciPy~\cite{2020SciPy-NMeth}. For experiments that use 3D Gaussian splats as input, we instead operate directly on point features using K-means clustering.

\subsection{Metric Definitions}
For easy comparison, we adopt the definition of the concavity metric from~\cite{wei2022approximate} as:
\begin{equation}
\resizebox{0.9\hsize}{!}{$
\mathrm{Concavity}(S)=
\max\big(
H(\partial S,\, \partial\,\mathrm{hull}(S)),
\; H(\mathrm{Vol}(S),\, \mathrm{Vol}(\mathrm{hull}(S)))
\big)
$}
\label{eq:supp:train_loss}
\end{equation}

\noindent where $H(\cdot)$ denotes the Hausdorff distance, $\partial S$ is the surface of $S$, and $\mathrm{Vol}(S)$ denotes points sampled in its interior. We uniformly sample 20{,}000 points per component to compute this metric. For reconstruction metric, we measure mean surface Chamfer distance between original shape and the union of convex hull, we also sample 20{,}000 points to calculate this metrics.

\section{Baseline Comparisons Details}
\label{sec:supp:baseline_comparison}
\vspace{3mm}

\paragraph{BSP-Net~\cite{chen2020bspnet}}
We use the official PyTorch codebase and the released checkpoint trained on ShapeNet. In addition, we train the network on the Objaverse subset used by our model. Following the official training protocol, we train the auto-encoding network (voxel input) for 8M steps at voxel resolutions 16 and 32, and for 16M steps at resolution 64 during Phase~0. We then train for another 16M steps at resolution 64 for Phase~1. We use the default network architecture and optimizer settings.

\paragraph{Cvx-Net~\cite{deng2020cvxnet}}
We use the official TensorFlow implementation. Since no pretrained checkpoint is available, we train the network separately on ShapeNet and on our Objaverse subset. The model takes 20 depth images as input. Following their original protocol, for each shape, we sample random camera positions uniformly on a sphere of radius 1.5 centered at the origin and render depth maps at a resolution of $224\times224$. We train the network for 1M steps.

\paragraph{VHACD~\cite{vhacd}}
We run VHACD using the implementation in the PyBullet\cite{coumans2016pybullet} package. All shapes are processed with a 1M-voxel approximation.
Like our approach, the algorithm allows the user to select a threshold to control the granularity of the decomposition; finer decomposition granularity generally improves quality metrics at the expense of generating more components.
For fair comparison, we follow a similar approach to~\cite{wei2022approximate} in selecting the threshold value and manually search for a suitable value that roughly results in the same granularity, i.e. the same number of resulting hulls.
For Table~\ref{tab:comparison} and Table~\ref{tab:ablation}, we use a fixed threshold of 0.01.

\paragraph{CoACD~\cite{wei2022approximate}}
We use their official implementation in our comparisons. 
Similar to VHACD, the algorithm requires specifying a threshold to control the granularity of the decomposition; finer decomposition granularity generally improves quality metrics at the expense of generating more components.
For fair comparison, we follow their evaluation protocol in selecting the threshold value and manually search for a suitable value that roughly results in the same granularity, i.e. the same number of resulting hulls.
For Table~\ref{tab:comparison} and Table~\ref{tab:ablation}, we use a fixed threshold of 0.10.

\section{Run time comparison with baseline methods}
\label{sec:supp:run_time_comparison}
Table ~\ref{tab:run_time} compares run time statistics with baseline methods. Our approach is somewhat faster than optimization-based methods and somewhat slower than feedforward learning model, but offers quality improvements over both.  

\begin{table}[t]
\centering
\small
\setlength{\tabcolsep}{3.5pt}
\renewcommand{\arraystretch}{1.1}
\begin{tabularx}{\linewidth}{c|CCCCC}
    \toprule
     & VHACD & CoACD & CVX & BSP & Ours  \\
    \midrule

    time (s) & 16.11 & 25.86 & 1.46 & 1.02 & 13.32\\

    memory (GB) & 0.5 & 5.5 & 1.78 & 1.38 & 2.61 \\
    GPU memory (GB) & - & - & 2.08 & 1.65& 3.61 \\
    model size (Mparams) & - & - & 44 & 39 & 106 \\
    \bottomrule
\end{tabularx}
\vspace*{-1em}
\caption{Per-shape computation cost on VHACD dataset}
\label{tab:run_time}
\vspace*{-1em}
\end{table}

\section{Additional Ablation with Alternative Training Objectives}
\label{sec:supp:additional_ablation}

We provide an additional ablation to validate the effectiveness of our loss function. Specifically, we ablate on our triplet-based contrastive learning objective. We compare it against a network trained on our feature embedding objective defined in Equation~\ref{eq:contrastive_opt_plain}, but ~\emph{without} using triplets. Concretely, we train the model with the following loss:

\begin{equation}
L = \tfrac{1}{2}\Bigl[-\,f(x)^{\top}f(p) \;+\; f(x)^{\top}f(n)\Bigr]
\label{eq:supp:alternative_loss}
\end{equation}

\begin{table}[t]
\centering
\small
\setlength{\tabcolsep}{3.5pt}
\renewcommand{\arraystretch}{1.1}
\begin{tabularx}{\linewidth}{c|CCC|CCC}
    \toprule
     & \multicolumn{3}{c|}{VHCD Dataset} & \multicolumn{3}{c}{Objaverse-Tiny} \\
     & \#comp & conc. & recon. & \#comp & conc. & recon. \\
    \midrule

    Alter. objective & 15.24& 0.1243 & 0.0260 & 27.36 & 0.1653 & 0.0303 \\
    \textbf{Ours} & \textbf{11.90} & \textbf{0.0973} & \textbf{0.0180} & \textbf{21.18} & \textbf{0.1257} & \textbf{0.0254} \\
    \bottomrule
\end{tabularx}
\caption{Quantitative ablation on alternative training objectives.}
\label{tab:supp:ablation}
\end{table}

As shown in Table~\ref{tab:supp:ablation}, the triplet-based contrastive loss used in our main method outperforms this alternative objective, demonstrating its effectiveness.

\section{More Results}
\label{sec:supp:more_result}

We present additional qualitative results across various granularities for further evaluation.
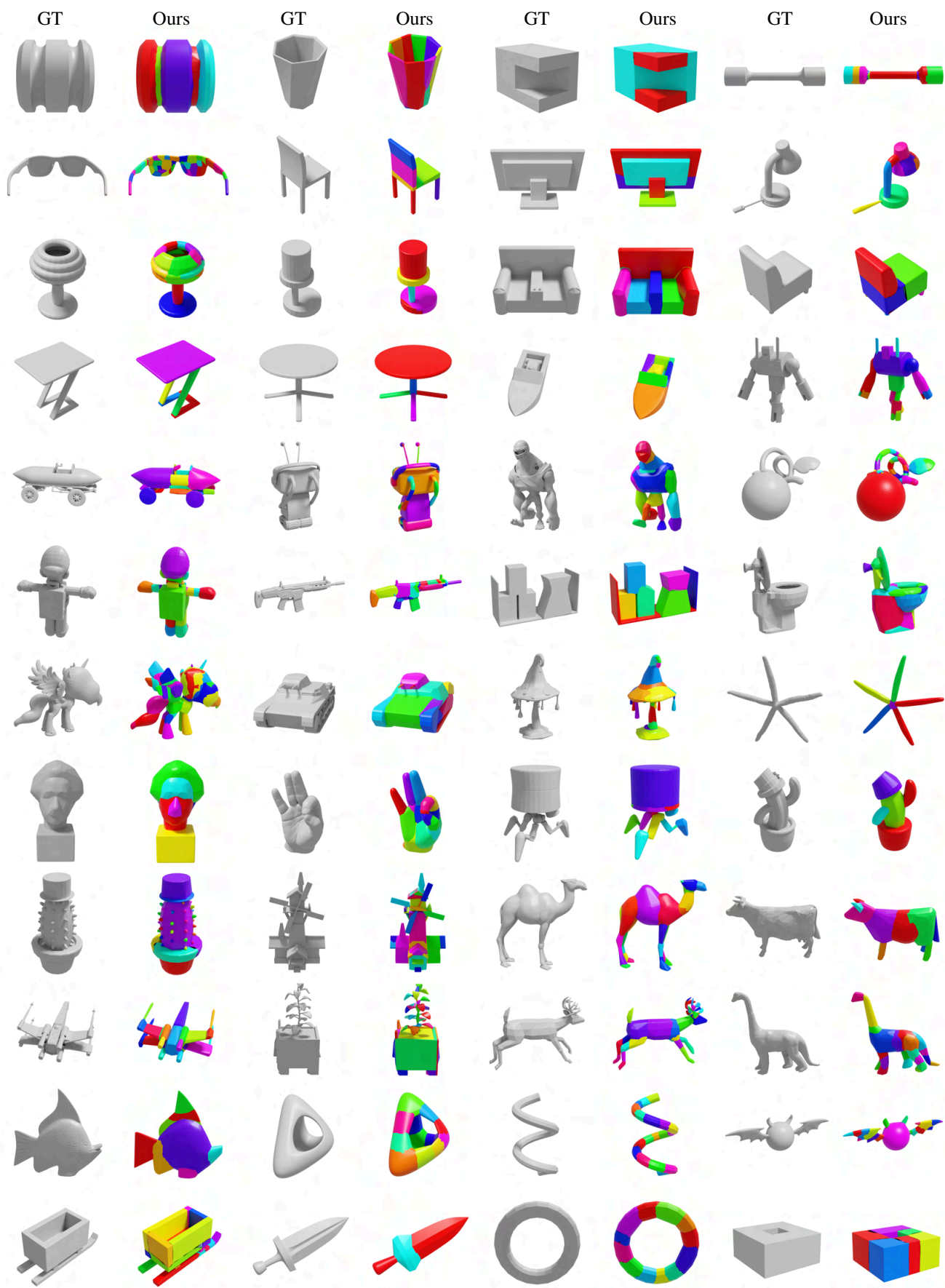
\begin{figure*}[h]
\centering
\begin{overpic}[width=1.0\linewidth]{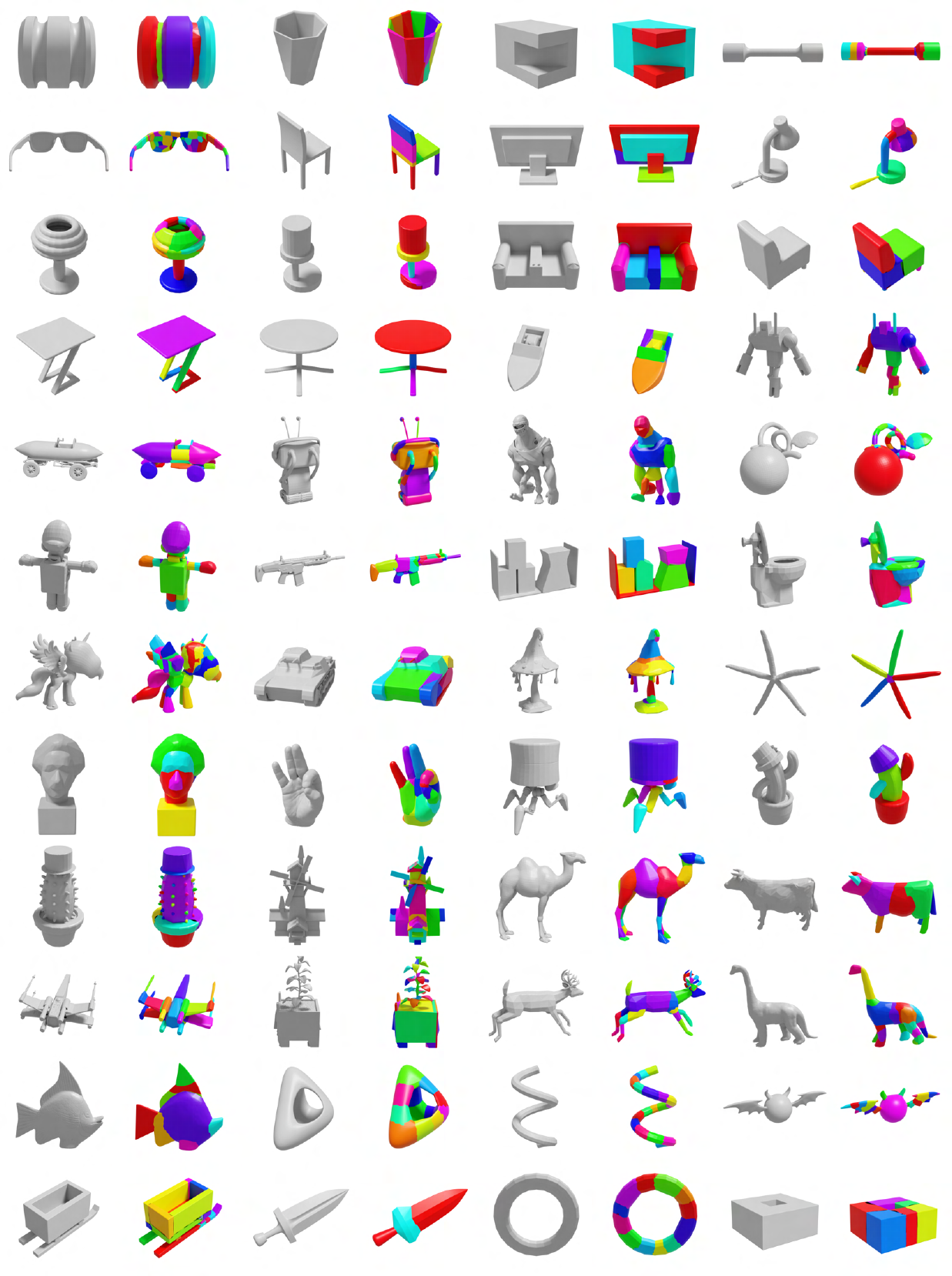}
    \put(3,100){GT}
    \put(12,100){Ours}
    \put(22,100){GT}
    \put(31,100){Ours}
    \put(41,100){GT}
    \put(50,100){Ours}
    \put(60,100){GT}
    \put(68,100){Ours}
\end{overpic}
\caption{Additional result. 
}
\label{fig:qualitative_comparison}
\end{figure*}

\end{document}